\documentclass[letterpaper, 10 pt, conference]{ieeeconf}  

\IEEEoverridecommandlockouts                              

\usepackage{graphicx} 
\graphicspath{{img/}}
\usepackage{amsmath} 
\usepackage{amssymb}  
\usepackage{paralist} 
\usepackage{booktabs} 
\usepackage{float} 
\usepackage{subcaption} 

\usepackage{algorithm}
\usepackage[noend]{algpseudocode}
\algnewcommand{\LineComment}[1]{\State \(\triangleright\) #1}

\makeatletter
\def\BState{\State\hskip-\ALG@thistlm}
\makeatother

\title{\LARGE \bf
Preparation of Papers for IEEE Sponsored Conferences \& Symposia*
}

\usepackage[backend=biber,style=ieee]{biblatex}
\bibliography{library}

\author{V. Strobel \and R. Meertens \and G.C.H.E. de Croon}
\date{\today}
\title{Efficient Global Indoor Localization for Micro Aerial Vehicles}

\begin{document}

\maketitle
\thispagestyle{empty}
\pagestyle{empty}

\begin{abstract}
  Indoor localization for autonomous micro aerial vehicles (MAVs)
  requires specific localization techniques, since the Global
  Positioning System (GPS) is usually not available. We present an
  efficient \emph{onboard} computer vision approach that estimates 2D
  positions of an MAV in real-time. This \emph{global} localization
  system does not suffer from error accumulation over time and uses a
  $k$-Nearest Neighbors ($k$-NN) algorithm to predict positions based
  on textons---small characteristic image patches that capture the
  texture of an environment. A particle filter aggregates the
  estimates and resolves positional ambiguities. To predict the
  performance of the approach in a given setting, we developed an
  evaluation technique that compares environments and identifies
  critical areas within them. We conducted flight tests to
  demonstrate the applicability of our approach. The algorithm has
a localization accuracy of approximately 0.6\,m on a 5\,m$\times$5\,m area at a runtime of 32 ms on
board of an MAV. Based on
  random sampling, its computational effort is scalable to different
  platforms, trading off speed and accuracy.
\end{abstract}


\section{INTRODUCTION}
\label{sec:introduction}

Accurate onboard localization is a key challenge for micro aerial
vehicles (MAV). In confined spaces, specific localization algorithms
are essential, since the Global Positioning System (GPS) is usually
not available. While light-weight MAVs could be employed in various indoor tasks,
they cannot fall back on standard localization approaches due to their
limited payload and processing power. To address this issue, this
paper presents an efficient indoor localization technique.

\begin{figure}
\includegraphics[width=\columnwidth]{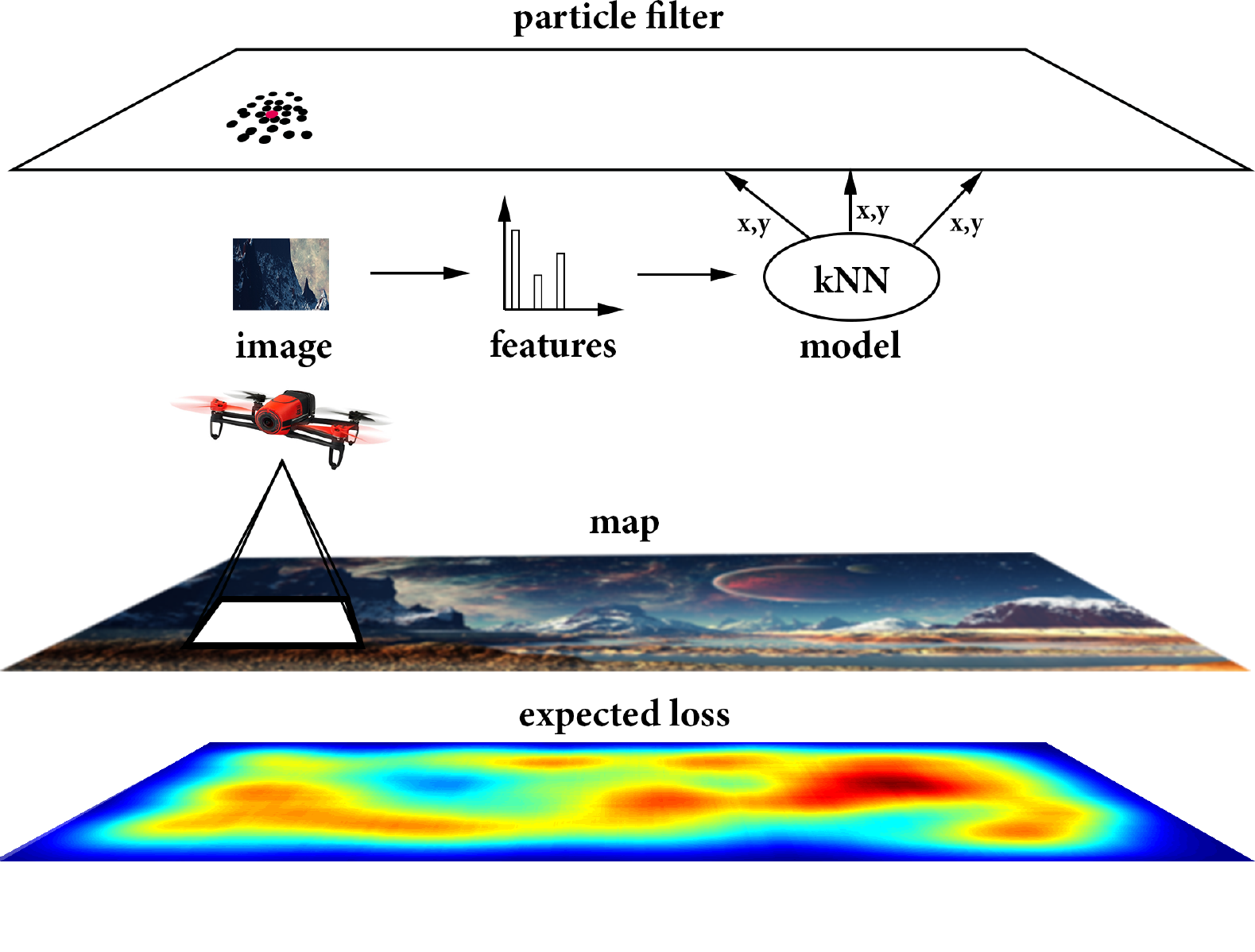}
\caption{{\label{fig:highleveloverview} The figure illustrates the
    presented system from a high-level perspective. A feature
    vector---the texton histogram---is extracted from the current
    camera image. The vector is forwarded to a machine learning model
    that uses a $k$-Nearest Neighbors algorithm to output $k$
    $x,y$-position estimates. These estimates are passed to a particle
    filter, which filters position estimates over time and outputs a
    final position estimate (red point). The expected loss shows
    regions in the map where a lower localization accuracy is
    expected. The average expected loss can be used as ``fitness
    value'' of a given map.%
  }}
\end{figure}

Our \textbf{contribution} is a machine learning-based indoor
localization system that runs onboard of an MAV  paving the way to
an autonomous system. In the presented approach, computational power is shifted to an offline training phase to achieve high-speed during live operation. In contrast to visual SLAM frameworks, this
project considers scenarios in which the environment is known beforehand or can even be
actively modified. The approach is based on the occurrence of
textons, which are small characteristic image patches. With textons as
image features and a $k$-Nearest Neighbors ($k$-NN) algorithm, we
obtain 2D positions in real-time within a known indoor environment. A
particle filter was developed that handles the estimates of the $k$-NN
algorithm and resolves positional ambiguities. We consider settings in which the MAV moves at an
approximately constant height, such that the estimation of height is
not necessary. In contrast to existing approaches that use \emph{active} sensors, the
developed approach only uses a \emph{passive} monocular
downward-looking camera. While carrying active sensors, such as
laser range finders, is too demanding for a light-weight MAV, onboard cameras
can typically be attached. Additionally, we developed a technique for evaluating the
suitability of a given environment for the presented algorithm. It
identifies critical areas and assigns a global loss value to an
environment. This allows for comparing different potential maps and
identifying regions with low expected localization accuracy.

The developed global localization system does not
suffer from error accumulation over time. Since it is intended to
further reduce the size of MAVs, lightweight and scalable position
estimation algorithms are needed. Onboard processing helps to reduce
errors and delays introduced by wireless communication, and ensures a
high versatility on the way to an autonomous system. The validity of
the approach is evaluated in flight experiments. An overview of the presented approach can be seen in
Figure~\ref{fig:highleveloverview}.

The remainder of this paper is structured as follows.
Section~\ref{sec:related} surveys existing indoor localization
approaches. In Section~\ref{sec:methods}, the developed texton-based
approach is presented and its components, the $k$-NN algorithm and the
particle filter, are introduced. Section \ref{sec:experiments}
describes the setup and results of the flight experiments. The results
are discussed in Section~\ref{sec:discussion} and we draw conclusions
in Section~\ref{sec:conclusion}.

\section{RELATED WORK }
\label{sec:related}

While a wide range of methods for indoor localization exists, we only
consider methods in this section that use the same technical and
conceptual setup---localization with a monocular camera.

One distinguishes two types of robot localization: local techniques
and global techniques~\cite{fox1999monte}. Local techniques need an
initial reference point and estimate coordinates based on the change
in position over time. Once they lost track, the position can
typically not be recovered. The approaches also suffer from ``drift''
since errors are accumulating over time. Global techniques are more
powerful and do not need an initial reference point. They can recover
when temporarily losing track and address the \emph{kidnapped robot
  problem}, in which a robot is carried to an arbitrary
location~\cite{engelson1992error}.

\subsection{Optical Flow}
\label{sec:opticalflow}

Optical flow algorithms estimate the apparent motion between
successive images. The most popular optical flow methods are gradient
based approaches and keypoint-based methods~\cite{chao2013survey}.
Optical flow methods belong to the class of \emph{local} localization
techniques and can only estimate the position relative to an initial
reference point. The approaches suffer from accumulating errors over
time and typically do not provide a means for correcting these errors.
Most approaches are computationally rather
complex~\cite{chao2013survey}.

\subsection{Fiducial Markers}
\label{sec:fiducialmarkers}

Fiducial markers have been used for UAV localization and
landing~\cite{eberli2011vision,bebop2015}. The markers encode
information by the spatial arrangement of black and white or colored
image patches. Their corners can be used for estimating the camera
pose at a high frequency.

An advantage of fiducial markers is their widespread use, leading to
technically mature and open-source libraries. A drawback
of the approach is that motion blur, which frequently occurs during
flight, can hinder the detection of
markers~\cite{albasiouny2015mean}. Furthermore, partial occlusion of
the markers through objects or shadows break the detection. Another downside is that markers might
be considered as visually unpleasant and may not fit into a product or
environmental design~\cite{chu2013halftone}.



\subsection{Homography Determination \& Keypoint Matching}
\label{sec:keypointmatching}

A standard approach for estimating camera pose is detecting and
describing keypoints of the current view and a reference
image~\cite{se2002global}, using algorithms such as Scale-invariant
feature transform (\textsc{Sift})~\cite{lowe1999object}, followed by
finding a homography---a perspective transformation---between both
keypoint sets. A keypoint is a salient image location described by a
feature vector. Depending on the algorithm, it is invariant to
different viewing angles and scaling.

This \emph{homography-based} approach is employed in frameworks for
visual Simultaneous Localization and Mapping (SLAM) but the pipeline
of feature detection, description, matching, and pose estimation is
computationally complex~\cite{kendall2015posenet}. While the approach
has been employed for global localization for UAVs, the required
processing power is still too high for small
MAVs~\cite{de2009design}.

%
%

\subsection{Convolutional Neural Networks}

Convolutional neural networks (CNNs) are a specialized machine learning
method for image processing~\cite{lecun1998gradient}. The supervised
method has outperformed other approaches in many computer vision
challenges~\cite{dosovitskiy2014discriminative}.  While their training
is usually time-consuming, predictions with CNNs often takes only few
milliseconds, shifting computational effort from the test phase to the
training phase. CNNs have been used as robust alternative for keypoint
detection and description if images were
perturbed~\cite{dosovitskiy2014discriminative} but needed more
computation time than \textsc{Sift}.

In recent work, \citeauthor{kendall2015posenet} present a framework
for regressing camera positions based on
CNNs~\cite{kendall2015posenet}. The approach is rather robust to
different lighting settings, motion blur, and varying camera
intrinsics. The approach predicts positions on a modern desktop
computer in short time.

\subsection{Texton-based Methods}
\label{sec:textonbasedapproaches}

Textons~\cite{varma2005statistical} are small characteristic image patches; their frequency in an
image can be used as image feature
vector. A texton histogram is obtained by extracting patches from an image and
comparing them to all textons in a ``texton dictionary''. The
frequency of the most similar texton is then incremented in the
histogram.

Texton histograms are flexible image features and their extraction
requires little processing time, which makes them suitable for MAV
on-board algorithms. The approach allows for adjusting the
computational effort by modifying the amount of extracted image
patches, resulting in a trade-off between accuracy and execution
frequency~\cite{de2012sub}. 

\citeauthor{de2009design}~\cite{de2009design} use textons
as image features to distinguish between three height classes of the
MAV during flight. Using a nearest neighbor classifier, their approach
achieves a height classification accuracy of approximately 78\,\% on a
hold-out test set. This enables a flapping-wing MAV to roughly hold
its height during an experiment. In another work,
\citeauthor{de2012appearance}~\cite{de2012appearance} introduce the
\emph{appearance variation cue}, which is based on textons, for
estimating the proximity to objects~\cite{de2012appearance}.
Using this method, the MAV achieves a high accuracy for collision
detection and can avoid obstacles in a $5\,m \times 5\,m$ office space.

\section{METHODS}
\label{sec:methods}

The pseudo code in Algorithm~\ref{alg:trexton_run} shows a high-level overview of the parts of the framework. Details are given in the following sections.

\begin{algorithm}
    \caption[High-level texton framework.]{High-level texton framework}
    \label{alg:trexton_run}
    \begin{algorithmic}[1]
      \State $t \gets 0$
      \State $\mathcal{X}_0 \gets$ \Call{init\_particles}{}
      \While {true}
      \State $t \gets t+1$ \State
      $I_t \gets$ \Call{Receive\_Img\_from\_Camera}{} \State
      $\mathcal{H}_t \gets$ \Call{Get\_Texton\_Histogram}{$I_t$}
      \State $\mathbf{z}_t \gets$
      \Call{$k$-NN}{$\mathcal{H}_t$}
      \State $\mathcal{X}_t \gets$
      \Call{Particle\_Filter}{$\mathcal{X}_{t-1}, \mathbf{z}_t$}
      \State $x_t, y_t \gets$ \Call{MAP\_Estimate}{$\mathcal{X}_t$}
      \EndWhile
\State \textbf{end}
    \end{algorithmic}
  \end{algorithm}  
  
  \subsection{Hardware and Software}
We decided to use the quadcopter \emph{Parrot Bebop Drone} as a prototype for all our
tests. The developed approach makes use of the bottom camera only,
which has a resolution of 640 $\times$ 480 pixels with a frequency of
30 frames per second.

\subsection{Dataset Generation}
A main idea of the presented method is to shift computational effort
to a pre-flight phase. Since the MAV will be used in a fixed
environment, the results of these pre-calculations can be employed
during the actual flight phase. Supervised machine learning methods
need a training set to find a mapping from features to target
values. In this first step, the goal is to label images with the
physical $x,y$-position of the UAV at the time of taking the image.

One possible way to create the data set is to align the images with
high-precision position estimates from a motion tracking system, which yields
high-quality training sets. Major disadvantages of the approach are that motion
tracking systems are usually expensive and time-consuming to move to
different environments.

As an alternative, we sought a low-budget and more flexible
solution. Out of the presented approaches in Section~\ref{sec:related},
the homography-based approach (Section~\ref{sec:keypointmatching})
promises the highest flexibility with a good accuracy but also
requires the most processing time. Since fast processing time is not
relevant during the pre-flight phase, the approach is well-suited for
the problem.
The required image dataset can be obtained by using images gathered
during manual flight or by recording images with a hand-held
camera. To get a hyperspatial image of the scene for creating a map,
the images from the dataset have to be stitched together.  With
certain software packages the images can be orthrectified by
estimating the most probable viewing angle based on the set of all
images. However, since a downward-looking camera is attached to the
UAV, most images will be roughly aligned with the z-axis, given slow
flight~\cite{blosch2010vision}. For the stitching process, we used the freeware software Microsoft Image Composite Editor (ICE)~\cite{ice}.
Keypoints of the current image and the map image are detected and
described using the \textsc{Sift} algorithm. This is followed by a
matching process, that identifies corresponding keypoints between both
images. These matches allow for finding a homography between both
images. For determining the $x, y$-position of the current image, its
center is projected on the reference image using the homography
matrix.

\subsection{Texton Dictionary Generation}
For learning a suitable texton dictionary for an environment, image patches
were clustered. The resulting cluster centers---the prototypes of the
clustering result---are the textons~\cite{varma2003texture}. The
clustering was performed with a Kohonen
network~\cite{kohonen1990self}. The first 100 images of each dataset were used to generate the dictionary. From each image, 1\,000 randomly selected image patches of
size $w \times h = 6 \times 6$ pixels were extracted, yielding
$N = 100\,000$ image patches in total that were clustered. For our approach, we also used the color
channels U and V from the camera to obtain color textons.

\subsection{Histogram Extraction}
The images from the preliminary dataset are converted to the final
training set that consists of texton histograms and $x,y$-values. To
extract histograms in the \emph{full sampling} setting, a small
window---or kernel---is convolved across the width and height of an
image and patches are extracted from all positions. Each patch is
compared with all textons in the dictionary and is labeled with the
nearest match based on Euclidean distance. The
histogram is normalized by dividing the number of cases in each bin by
the total number of extracted patches, to yield the relative frequency
of each texton.

The convolution is a time-consuming step, since all possible
combinations of width and height are considered:
$(640 - w + 1) \cdot (480 - h + 1) = 301\,625$ samples are
extracted. To speed up the time requirements of the histogram
extraction step, the kernel can be applied only to randomly sampled
image position instead~\cite{de2012sub}. This sampling step speeds up
the creation of the histograms and permits a trade-off between speed
and accuracy.
The random sampling step introduces random effects into the
approach. Therefore, for generating the training dataset, no random
sampling was used to obtain high-quality feature vectors.

\subsection{$k$-Nearest Neighbors ($k$-NN) algorithm}
\label{sec:knn}

The $k$-Nearest Neighbors ($k$-NN) algorithm is the ``machine
learning-core'' of the developed algorithm. Taking a texton histogram
as input, the algorithm measures the Euclidean distance of this
histogram to all histograms in the training dataset and outputs the
$k$ most similar training histograms and the corresponding
$x,y$-positions.

While the $k$-NN algorithm is one of the simplest machine learning
algorithms, it offers several advantages~\cite{kordos2010we}: it is
non-parametric, allowing for the modeling of arbitrary
distributions. Its capability to output multiple predictions enables
neat integration with the developed particle filter. Additionally, $k$-NN
regression often outperforms more sophisticated
algorithms~\cite{knn}. A frequent point of criticism is its increasing
computational complexity with an increasing size of the training
dataset. While the used training datasets consisted of fewer than 1000
images, resulting in short prediction times (see also Figure~\ref{fig:freqhist}), time complexity can be
reduced by storing and searching the training examples in an efficient
manner, for example, with tree structures~\cite{bhatia2010survey}.


\subsection{Filtering}
\label{sec:filtering}

Computer vision-based estimations are often noisy or ambiguous. Texton
histograms obtained during flight will not perfectly match the ones in
the training dataset: blur, lighting settings, viewing angles, and,
other variables change the shape of the histograms.

A popular filter choice is the Kalman filter. However, the Kalman filter is not able to
represent multimodal probability
distributions. This makes it rather
unsuitable for the presented \emph{global} localization approach. The
``naive'' $k$-NN regression calculates the mean of the $k$ outputs and
forwards this value to the Kalman Filter.  We decided to use a more sophisticated method to capture
\emph{multimodal distributions}. Given an adequate measurement model,
a general Bayesian filter can simultaneously maintain multiple
possible locations and resolve the ambiguity as soon as one location
can be favored. In this case, the predictions
of the $k$ neighbors can be directly fed into the filter without
averaging them first. However, a general Bayesian filter is computationally
intractable. Therefore, a variant based on random sampling was used: the
particle filter. While its computational complexity is still high
compared to a Kalman filter, one can modify the amount of particles
to trade off speed and accuracy and adapt the computational payload to
the used processor.

The weighted particles are a discrete approximation of the probability
density function ($pdf$) of the state vector ($x,y$-position of the
MAV). Estimating the filtered position of the MAV can be described as
$p(X_t \mid Z_t)$, where $X_t$ is the state vector at time $t$ and
$Z_t = \mathbf{z}_1, ..., \mathbf{z}_t$ are all outputs of the $k$-NN
algorithm up to time $t$, with each $\mathbf{z}_i$ representing the
$k$ $x,y$-outputs of the algorithm at time $i$.

The used particle filter is initialized with particles at
random $x, y$-positions. To incorporate the measurement noise for each
of the $k$ estimates from the $k$-NN algorithm, we developed a
two-dimensional Gaussian Mixture Model (GMM) as measurement model.
The GMM is parameterized by the
variances~$\Sigma^{[j]}, j \in \{1, \ldots, k\}$ that are dependent on the
rank $j$ of the prediction of the $k$-NN algorithm (for example,
$j = 2$ is the second nearest neighbor).
The variance matrix $\Sigma^{[j]}$ specifies the variances of the
deviations in $x$-direction and $y$-direction and the correlation
$\rho$ between the deviations.
The values for $\Sigma^{[j]}$ were determined by calculating the
variance-covariance matrix for the difference between the ground truth
$T$ from the motion tracking system and the predictions $P_j$ of the $k$-NN algorithm:
$\Sigma^{[j]} := \text{Var}(T-P_j)$.

In contrast to the measurement model, the used \emph{motion model} is
simple. It is solely based on Gaussian process noise and does not
consider velocity estimates, headings, or control inputs. Its mean and
variance are dependent on the expected velocity of the MAV. We used
the forward difference $T_t - T_{t-1}$ to estimate the average
movement and its variance-covariance matrix $\Sigma_{\text{process}}$
between timesteps $t$ and $t-1$. 

The algorithm of the developed particle filter is presented in the
pseudo code in Algorithm~\ref{alg:particle_filter}. In the pseudo
code, $\mathcal{X}$ is the list of particles, $f$ the two-dimensional
Gaussian probability density function, $z_t^{[i]}$ the $i$th neighbor
from the $k$NN prediction, $x_t^{[m]}$ the $m$th particle at time $t$,
and $w_t^{[m]}$ its corresponding weight.
\begin{algorithm}
\caption[Particle filter update]{Particle filter update}
\label{alg:particle_filter}
\begin{algorithmic}[1]
  \Procedure{Particle\_Filter}{$\mathcal{X}_{t-1}, z_t$}
    \LineComment Initialize particle list
  \State $\mathcal{X}_{\text{temp}} := \varnothing$
  \For{$m = 1$ to $M$}
  \LineComment{Add random process noise (motion model)}
  \State $x_t^{[m]} \gets x_t^{[m]} + \mathcal{N}(0, \Sigma_{\text{process}})$
  \LineComment{Iterate over $k$-NN preds (measurement model)}
  \State $w \gets 0$
  \For{$i = 1$ to $k$}
  \LineComment{Gaussian Mixture Model}
  \State $w \gets w + f(z_t^{[i]} ; x_t^{[m]}, \Sigma_{\text{measurement}}^{[i]})$ 
       \EndFor
  \State $\mathcal{X}_{\text{temp}} := \mathcal{X}_{\text{temp}} \cup  (x_t^{[m]}, w)$
  \EndFor
  \LineComment{Importance resampling}
  \State $\mathcal{X}_t \gets$ \Call{Resampling\_Wheel}{$\mathcal{X}_{\text{temp}}$}
  \State \Return $\mathcal{X}_t$
  \EndProcedure
\end{algorithmic}
\end{algorithm}
The ``resampling wheel''~\cite{thrun} performs the importance
resampling step. 

With the GMM, the information of all $k$ neighbors can be used,
yielding a possibly multimodal distribution. While a multimodal
distribution allows for keeping track of several possible positions,
certain subsystems---for example a control loop---often need
\emph{one} point estimate. Using a weighted average of the particles
would again introduce the problem that it could fall into a low
density region (an unlikely position). Instead, we used a maximum a
posteriori (MAP) estimate, as described by
\citeauthor{driessen2008map}~\cite{driessen2008map}.

The estimation of \emph{uncertainty} was modeled using the spread of the particles---as
expressed by their variance in $x$-direction and $y$-direction.

\subsection{Map evaluation}

The performance of the developed method depends on the environment: a
texture-rich environment without repeating patterns will be better
suited than a texture-poor environment. Ideally, one would like to
know if the algorithm will work in a given environment. Therefore, we
propose an evaluation scheme that can compare different environments
and areas within an environment. This scheme assigns a global fitness
value or global loss value to a ``map''---expressed as dataset
$\mathcal{D}$ consisting of $N$ texton histograms $h_i$ and
corresponding $x,y$-coordinates $\text{pos}_i = (x_i, y_i)$. The
fitness value is intended to be proportional to the accuracy that can
be expected when using this dataset as training set for the developed
localization algorithm. The scheme allows for inspecting the dataset
and detecting regions within the map that are responsible for the
overall fitness value.

The idea behind the global loss function $L$ is that histograms $h_i$
and $h_j$ in closeby areas should be similar and the similarity should
decrease with increasing distance of the corresponding
$x,y$-coordinates $\text{pos}_i$ and $\text{pos}_j$. Therefore, the
approach is based on the difference between \emph{actual} and
\emph{ideal} texton histogram similarities in a dataset. The ideal
texton similarity distribution is modeled as a two-dimensional
Gaussian distribution around each $x,y$-position in the dataset. Using this idea, a histogram is
compared to all others by comparing expected similarities to actual
similarities. This results in a loss value per sample of the dataset
(local loss). Applying the algorithm to each sample in the dataset
yields the global loss of a dataset.

The method uses the cosine similarity ($CS$) to compare histograms:
\begin{align*}
CS(h_i, h_j) = \frac{h_i^Th_j}{||h_i||\,||h_j||}
\end{align*}
The cosine similarity has the convenient property that its values are
bounded between $-1$ and $1$. In the present case, since the elements
of the histograms are non-negative, it is even bounded between $0$ and
$1$. Let the function $f$ describe the non-normalized one-dimensional
Gaussian probability density function:
\begin{align*}
  f(x; \mu, \sigma) = e^{- \frac{(x - \mu)^2}{2 \sigma ^ 2}}  
\end{align*}
Since we assume that the ideal similarity in $x$-position is
independent of the $y$-position, the ideal two-dimensional similarity
function $d_e(\text{pos}_i, \text{pos}_j; \Sigma)$ can be modeled as
the product of the respective one-dimensional function~$f$:
\begin{align*}
d_e(\text{pos}_i, \text{pos}_j; \Sigma) = f(x_i; x_j, \sigma_x) \cdot f(y_i;
y_j, \sigma_y)
\end{align*}
This function is also bounded between $0$ and $1$, which makes the
functions $d_e$ and $CS$---ideal similarity and actual
similarity---easily comparable. In summary, we propose the following
global loss function ($L$) for evaluating a given dataset
($\mathcal{D}$):
\begin{align*}
  L(\mathcal{D}) &= \frac{1}{N^2}\sum_{i = 1}^{N} \sum_{j = 1}^{N}
                   CS(h_i, h_j) - f(x_i; x_j, \sigma_x) \cdot f(y_i; y_j, \sigma_y)                  
\end{align*}
The simple difference---in contrast to least absolute deviations or
least square errors---ensures that similarities that are \emph{less}
similar than the ideal similarity \emph{reduce} the loss. Therefore, a
high variation in texture is always seen as ``positive''. The
variances $\sigma_x$ and $\sigma_y$ specify the dimension of the
region, where similar histograms are desired. The lower their value,
the more focused the ideal similarity will be, requiring a high
texture variety for getting a low loss value. A high value might
overestimate the suitability of a dataset. While the approach is
relatively robust to the choice of the parameter values, we still need
to find a heuristic for suitable values.

\begin{figure}[h]
  \centering
  \includegraphics[width=0.5\textwidth]{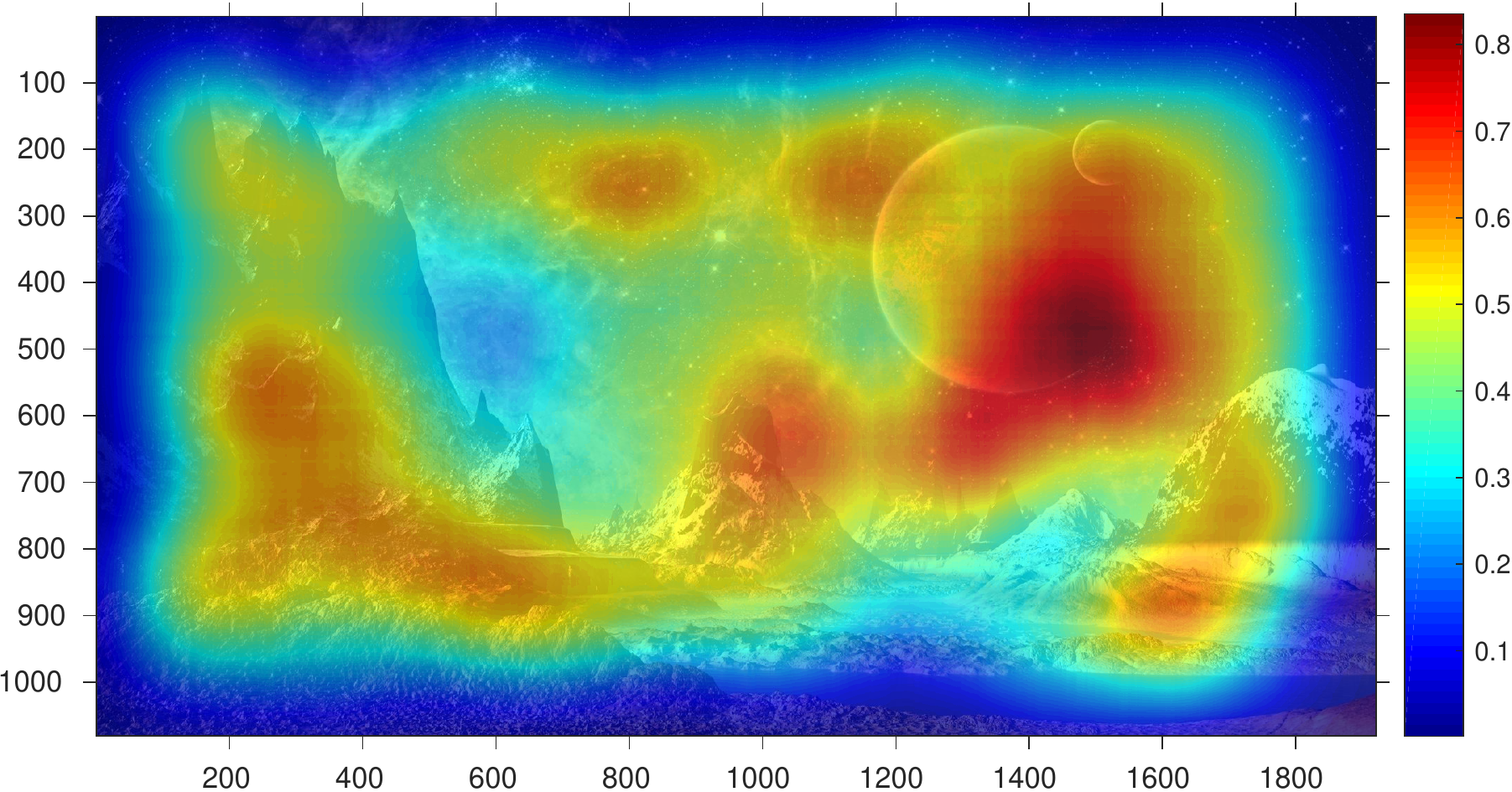}
  \caption[Global loss of a map.]{The figure shows the loss of a map: the regions that did
    not follow the ideal similarity pattern are displayed in red. For
    the visualization, the loss values per sample in the dataset were
    smoothed with a Gaussian filter. This assigns a loss value to each
    $x,y$-position of the map.}
  \label{fig:globalloss}
\end{figure}

\section{ANALYSIS}
\label{sec:experiments}

In the experiments, the MAV was guided along flight plans using the
motion tracking systen. If not otherwise stated, we used the following
default values for the parameters in our framework.
\begin{itemize}
\item number of samples in the histogram extraction step: 400
\item number of textons in the dictionary: 20
\item number of particles of the particle filter: 50
\item number of histograms / images in the training set: 800
\item number of histograms / images in the test set: 415
\item number of neighbors in the $k$-NN algorithm: 5
\end{itemize}
Map-dependent texton dictionaries were used and created by
conducting an initial flight over the respective maps.

\subsection{Baseline: Homography-based Approach}
\label{sec:siftvsoptitrack}

\begin{figure}[H]
\begin{center}
\includegraphics[width=0.7\columnwidth]{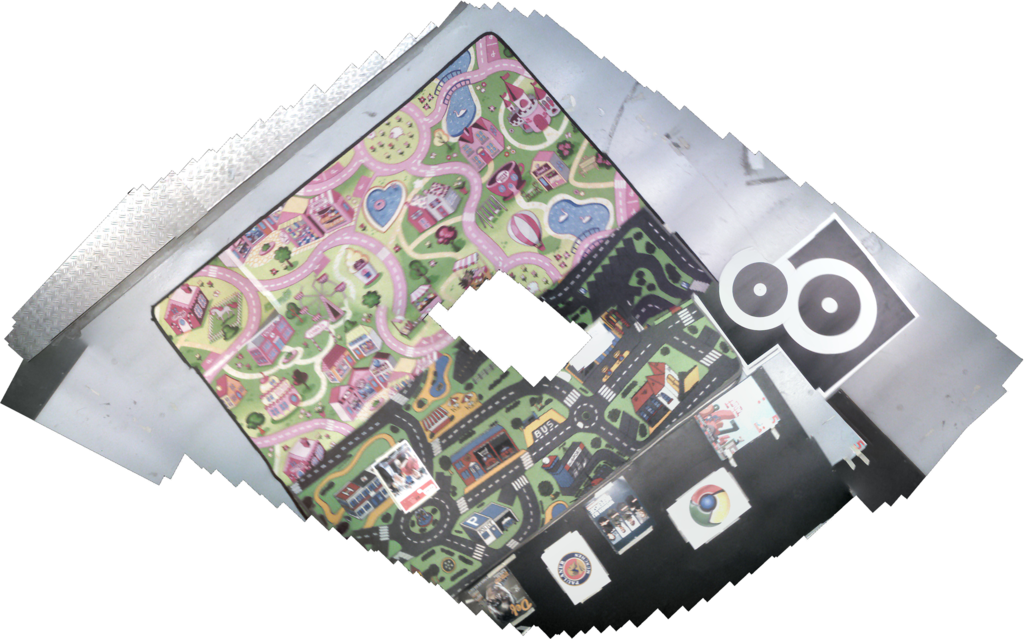}
\caption[Hyperspatial image of the map.]{{\label{fig:mapexp}The created map (size: approximately $5 \times 5$ meters) that was stitched
    together using 800 images. The ``non-mapped'' area in the center
    of the image is a result of the set flight path. An image
    distortion can be seen at the right-hand side, where the landing
    spot appears twice, while in reality, only one circle was
    visible.%
  }}
\end{center}
\end{figure}

To find a baseline for our approach and to provide a homography-based
training set, we used the homography-based approach to estimate
$x,y$-coordinates in the same environment and based on the same images
as the texton-based framework. The required hyperspatial image
(Figure~\ref{fig:mapexp}) of the environment was stitched together
using 800 images and the software Microsoft ICE.

We estimated the $x,y$-coordinates of the 415 test images using the
homography-based approach and compared the
predictions to the ground truth. The predictions were not
filtered. The results can be found in the following table.


\begin{table}[H]
  \centering
  \begin{tabular}{lrrr}
    \toprule
    & x-position & y-position\\
    \midrule
    Error in cm & 31 & 59\\
    STD in cm & 68 & 77\\
    \bottomrule
  \end{tabular}
  \label{tab:homoerror}
\end{table}

\subsection{Training Set based on Motion Tracking System}
\label{sec:experiment-real}

In this experiment, the position estimates were calculated on board of
the MAV using the texton-based approach with the particle filter. The
Euclidean distances between the estimates of the motion tracking
system and the texton-based approach were measured in $x$-direction
and $y$-direction.

The training dataset was composed of 800 texton histograms with
corresponding $x,y$-coordinates that were obtained from the motion
tracking system. The images were recorded in a $5 \times 5$ meter area
at a height of approximately one meter in a time span of one hour
before the experiment to keep environmental factors roughly the same.

The results can be found in the following table. They are based on
415 images, which corresponds to a flight time of approximately 35
seconds.
\begin{table}[h]
\centering
  \begin{tabular}{lrrr}
    \toprule
    & x-position & y-position\\
    \midrule
    Error in cm & 61 & 59\\
    STD in cm & 39 & 39\\
    \bottomrule
  \end{tabular}
\end{table}

\subsection{Training Set based on Homography-finding Method}
\label{sec:traininghomo}

In this experiment, the training dataset was created by estimating the
$x,y$-positions of the 800 training images using the
homography-finding method from the previous section and the same
hyperspatial image. Apart from that, the settings are the same as in the previous experiment.
\begin{table}[h]
\centering
  \begin{tabular}{lrrr}
    \toprule
    & x-position & y-position\\
    \midrule
    Error in cm & 54 & 97\\
    STD in cm & 41 & 61\\
    \bottomrule
  \end{tabular}
\end{table}

\subsection{Triggered Landing}
\label{sec:triggered}

For the triggered landing experiment, the MAV was guided along random
flight paths, which covered a $5 \times 5$ meter area; during
navigation, the MAV was programmed to land as soon as its position
estimates were in a ``landing zone'': an $x,y$-position with a
specified radius $r$. A safety criterion was introduced such that the
landing is only performed if the standard deviations of the particles
in $x$-direction and $y$-direction are below thresholds $\theta_x$ and
$\theta_y$. We set the parameters to $\theta_x = \theta_y = 60$\,cm.
The $x,y$-coordinate of the circle was specified in the flight plan;
the radius was set to $r = 60$\,cm. We performed six triggered
landings; after each landing, the $x,y$-center of the zone was
randomly set to another position in the map. For the texton framework,
the same training set as in Experiment~\ref{sec:experiment-real} was
used.

Four out of six landings were correctly performed in the landing
area. The distances of the two outliers were 14\,cm and 18\,cm,
measured as distance to the circumference of the landing area.

\subsection{Speed versus Accuracy Trade-Off}
\label{sec:speedvsacc}


Adapting the frequency of the main loop of the developed approach to
make it suitable for different platforms with varying processing power
is one of its core parts.

\begin{figure*}
  \centering
  \begin{subfigure}[b]{0.33\textwidth}
  \includegraphics[width=1\textwidth]{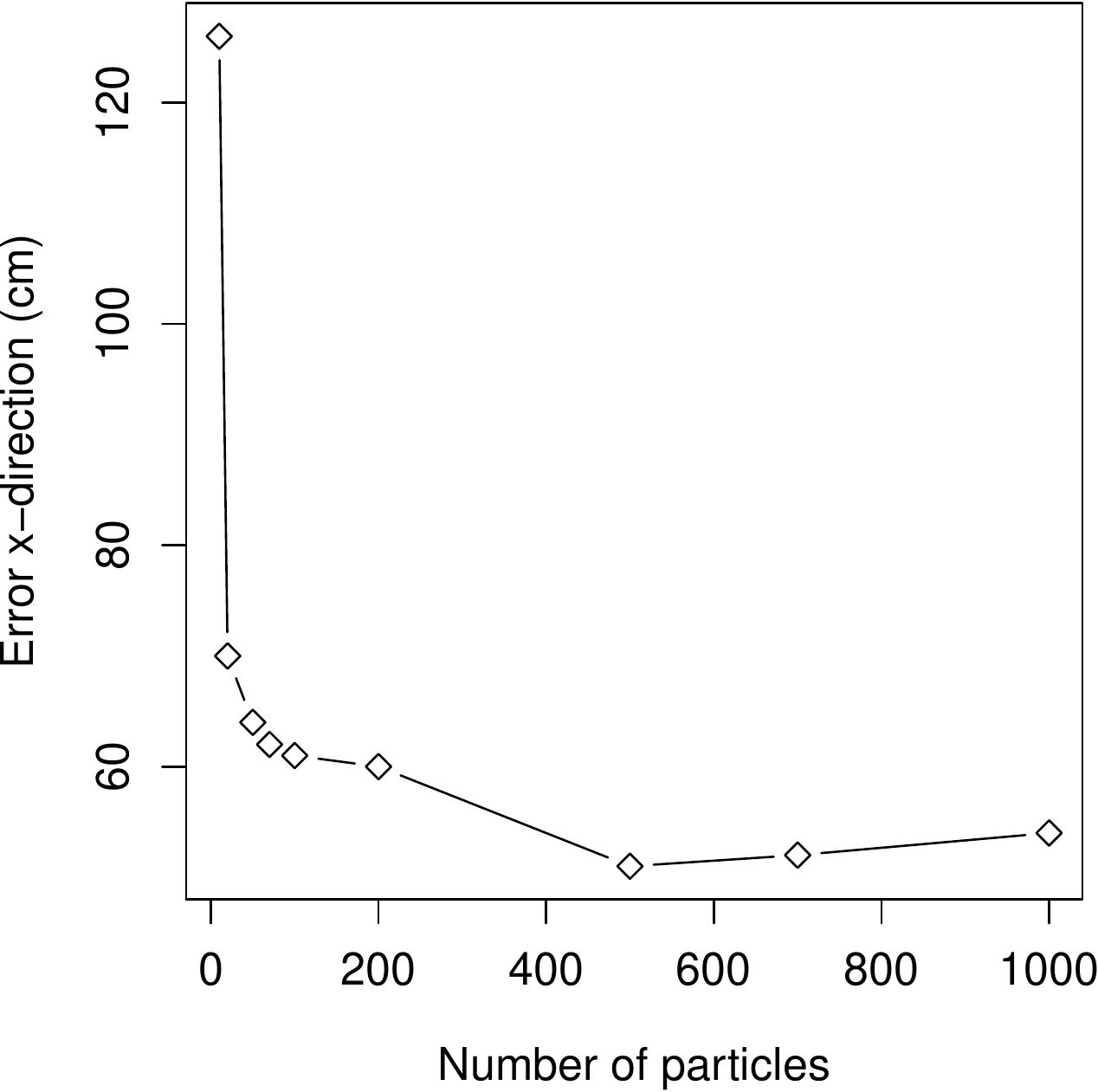}
  \end{subfigure}%
~
  \begin{subfigure}[b]{0.33\textwidth}
  \includegraphics[width=1\textwidth]{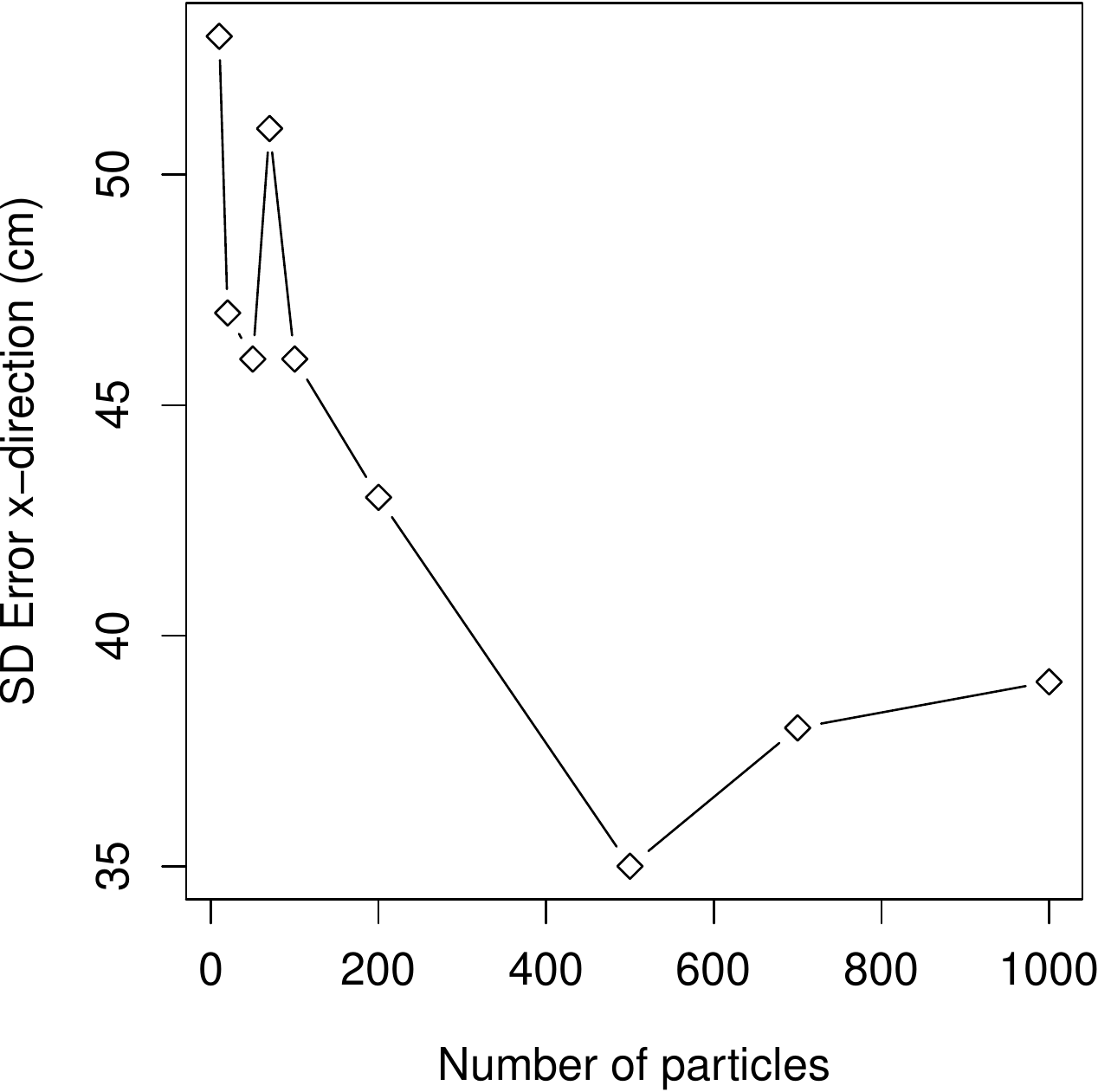}
  \end{subfigure}%
~
  \begin{subfigure}[b]{0.33\textwidth}
  \includegraphics[width=1\textwidth]{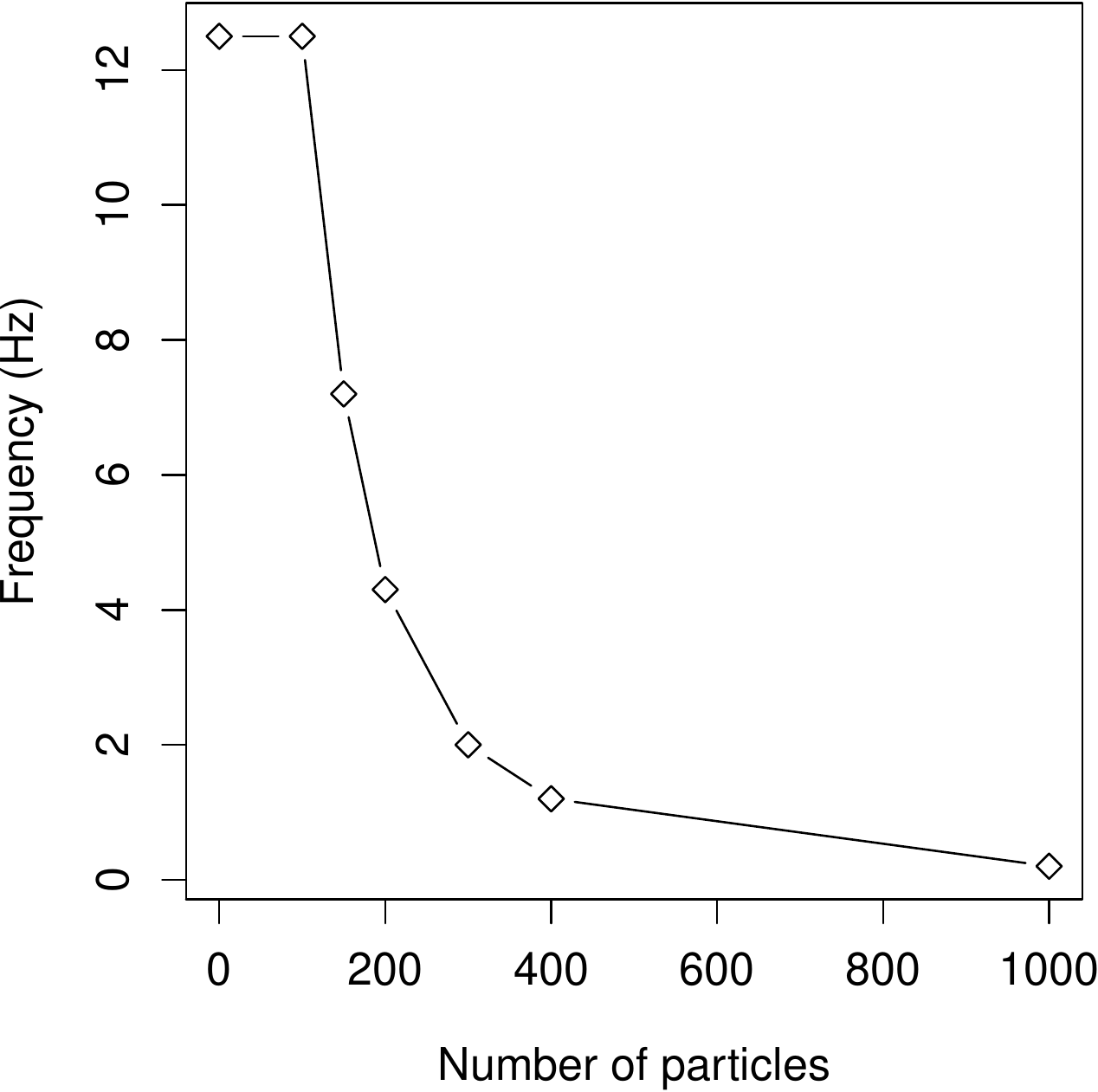}
  \end{subfigure}
  \caption{Speed versus accuracy trade-off in $x$-direction as a function of the number of used particles.}
  \label{fig:tradepart}
\end{figure*}

\begin{figure*}
  \centering
  \begin{subfigure}[b]{0.33\textwidth}
  \includegraphics[width=1\textwidth]{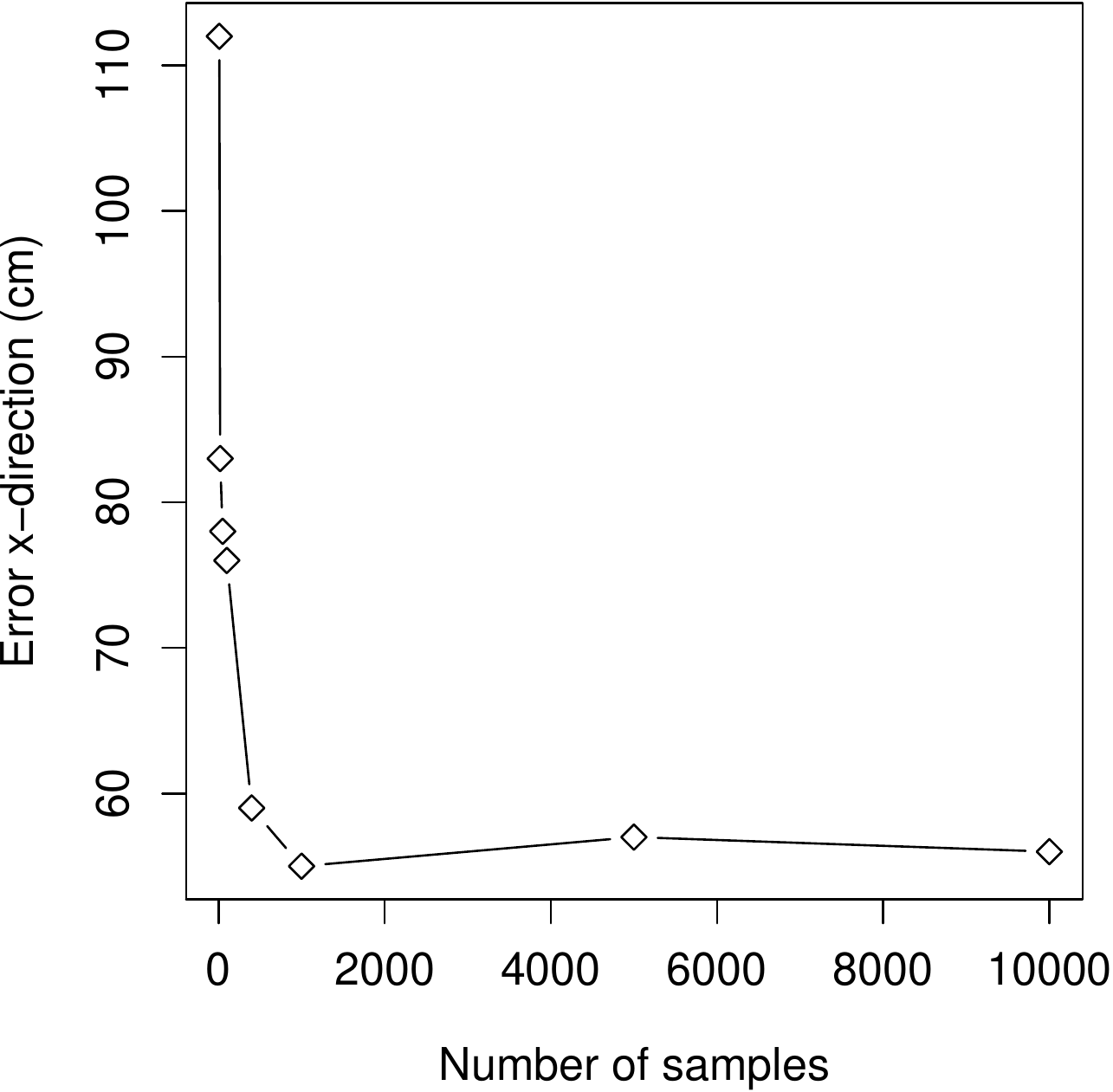}
  \end{subfigure}%
~
  \begin{subfigure}[b]{0.33\textwidth}
  \includegraphics[width=1\textwidth]{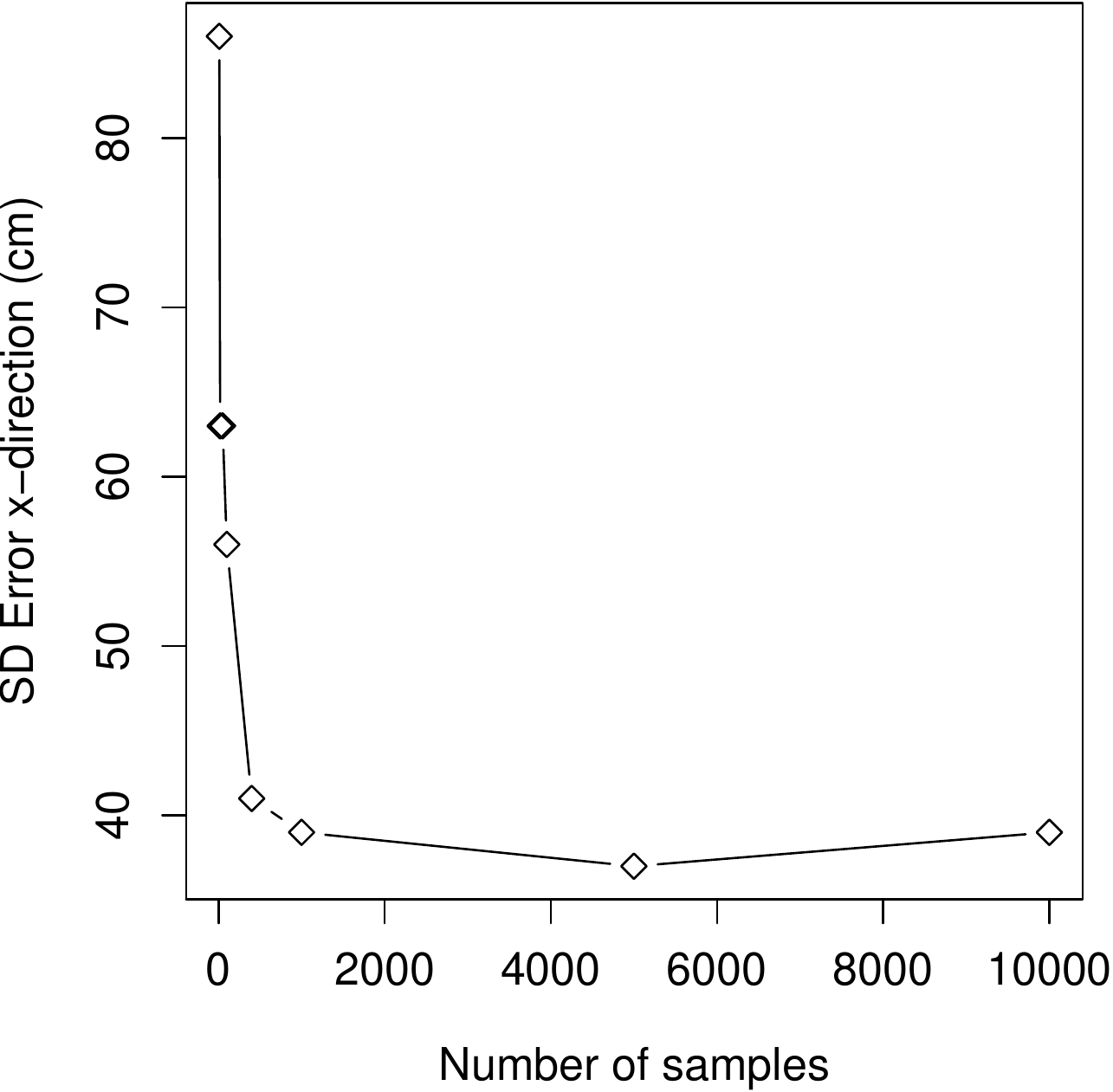}
  \end{subfigure}%
~
  \begin{subfigure}[b]{0.33\textwidth}
  \includegraphics[width=1\textwidth]{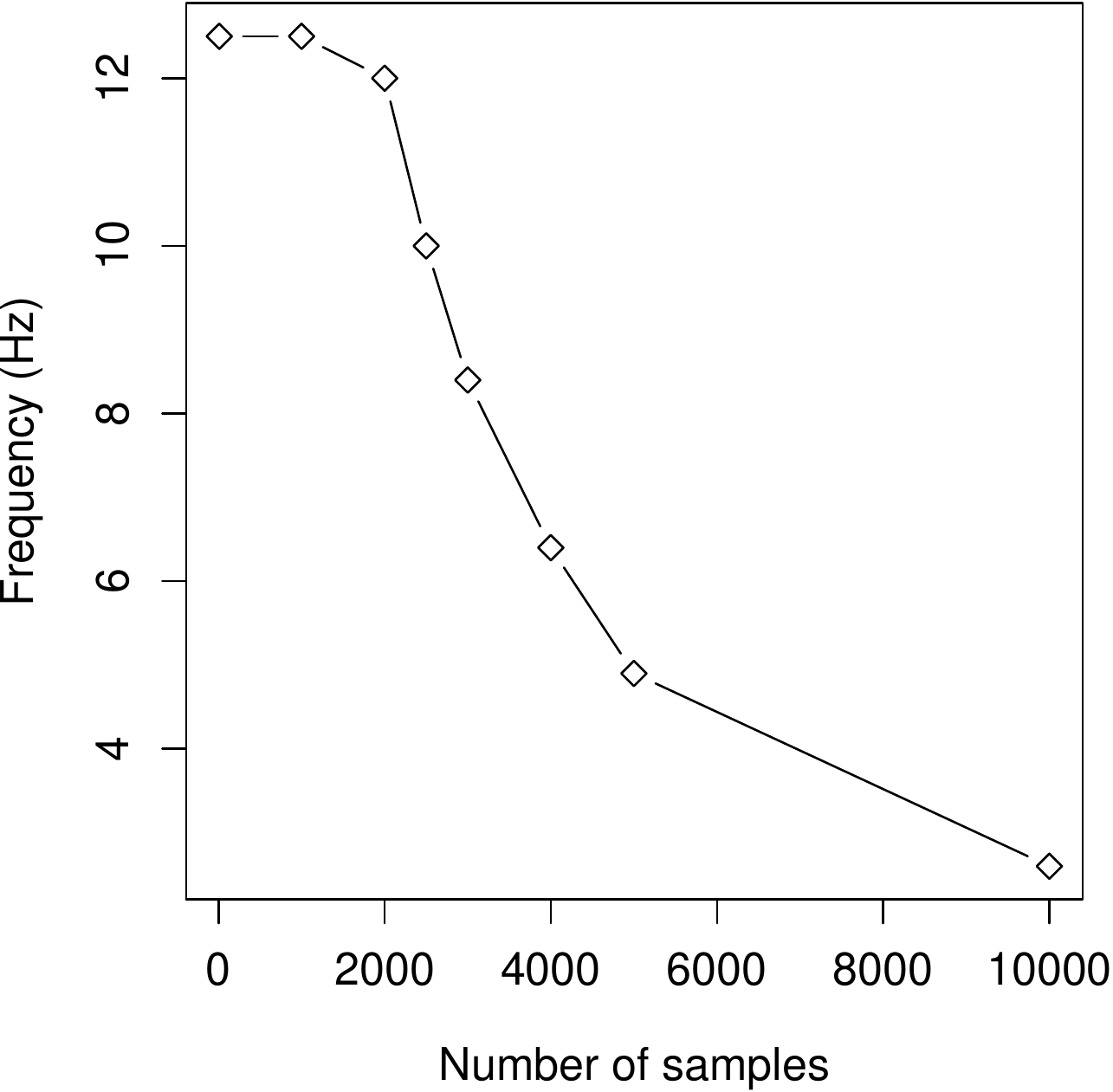}
  \end{subfigure}
  \caption{Speed versus accuracy trade-off in $x$-direction as a function of the number of used samples in the histogram extraction step.}
  \label{fig:tradesamples}
\end{figure*}

Figures~\ref{fig:tradepart} and~\ref{fig:tradesamples} show the speed
versus accuracy trade-off as a function of the used particles and of
the used samples in the histogram extraction step, respectively. As a
reference, the frequency using \emph{full sampling} in the histogram
extraction step was 0.1\,Hz. The above stated \emph{default} values
were used for the \emph{ceteris paribus} assumption, when varying the
parameters.

While the bottom camera of the Parrot Bebop Drone has a frequency of
30\,Hz, the Paparazzi software currently only receives the images with
a frequency of 12.5\,Hz. Therefore, the maximum achievable frequency
without further image processing is 12.5\,Hz, which is the baseline
for the conducted experiments.

\begin{figure*}
  \centering
\includegraphics[width=0.33\textwidth]{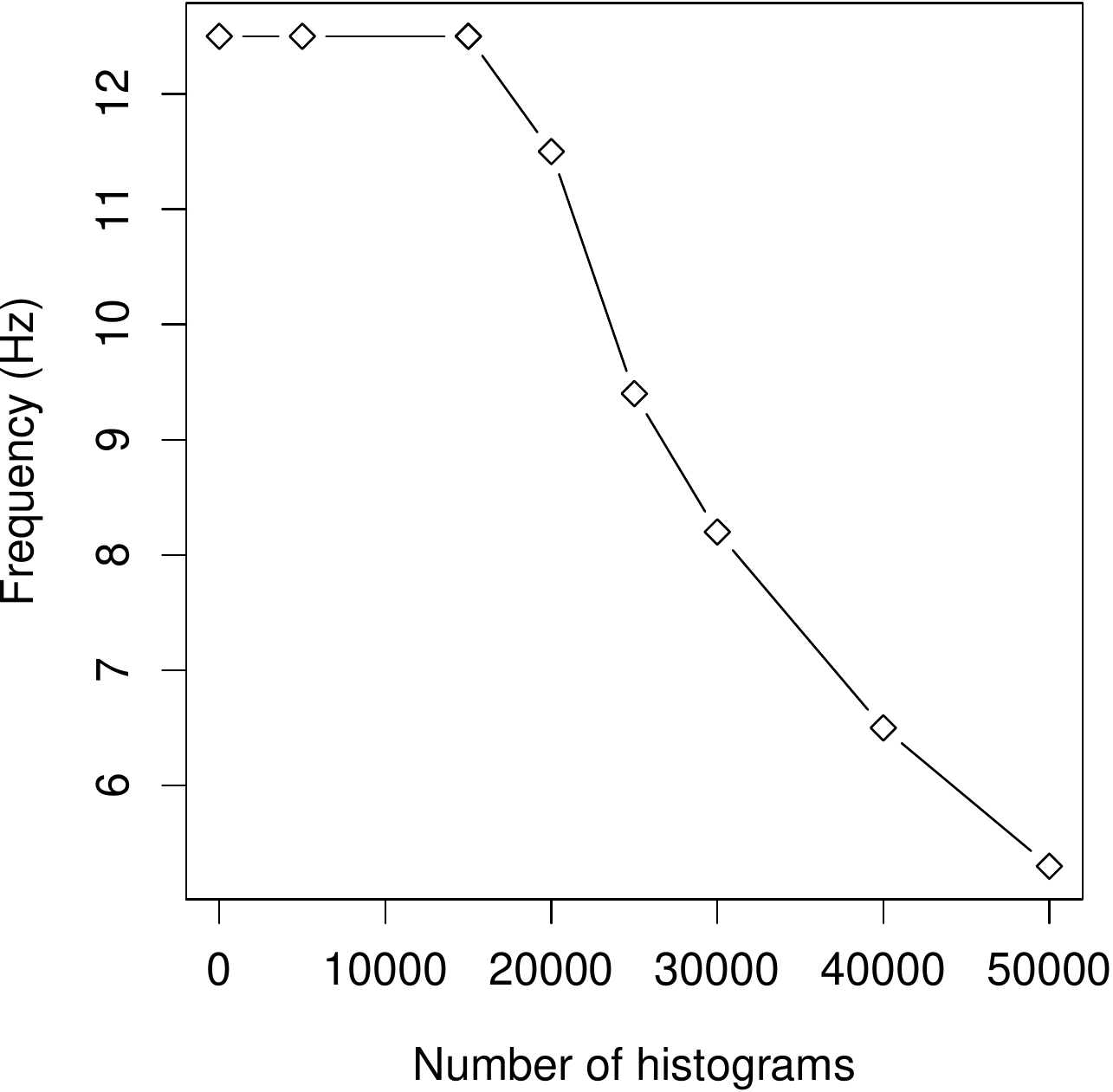}
  \caption[Frequency as a function of the number of histograms.]{Frequency of the main loop as a function of the number of
    histograms in the training set.}
  \label{fig:freqhist}
\end{figure*}

Figure~\ref{fig:freqhist} illustrates the frequency as a function of
the used histograms in the $k$-NN algorithm. We did not compare the
frequency to the distance between ground truth and the predictions,
since our training dataset did not contain more than 800 histograms.

After having received the image, the processing time of the presented
algorithm using the \emph{default} parameter values is 32\,ms, which
includes the histogram extraction (16\,ms) as well as the $k$-NN
predictions, the filtering and the output of the best $x,y$-coordinate
(16\,ms).

\section{DISCUSSION}
\label{sec:discussion}

The flight tests show initial evidence for the real-world suitability of the method. It yields slightly less accurate results than the unfiltered homography-finding method. While we did not test the frequency of the homography-based approach on board of an MAV, on a desktop computer,
it took 200 ms per image. Therefore, the developed algorithm runs at a much higher
frequency. The training set generation based on the homography method yielded higher
errors in the flight test. Filtering the estimates of the homography-method first could
improve the accuracy. The triggered landing (Experiment 4.4.4) showed good accuracy:
while most landings were triggered inside the landing zone, two out of the six landings
were outliers. However, their distance to the landing area were rather small, with an
average distance of 16 cm.

The experiments addressing the ``Speed versus accuracy trade-of'' show that
with an increasing accuracy of the approach, the frequency of the algorithm decreases.
However, the errors reach a plateau after which no large improvements can be expected
at the lower end of parameter ranges. By optimizing the parameters, one can obtain
localization errors below 50\,cm with the developed approach.

While we compared the settings of different parameters, there are no generally optimal
parameters for the presented framework: setting the number of textons, the number
of images patches, or the number of neighbors is dependent on the environment and
the size of the training dataset. The parameters have to be adapted to the particular
environment.

Despite the overall promising results of our localization algorithm, we noticed drawbacks
during the flight tests and identified several directions for future research that are described
in what follows.
The accuracy could be further improved by combining the presented
global localization technique with a local technique. To this end, odometry estimates using
optical flow or the inclusion of data from the inertial measurement unit (IMU) could be suitable.

Our current implementation assumes constant height up to few centimeters and only small
rotations of the MAV. While a quadroter can move in every direction without performing
yaw movements, other MAVs or the use of the front camera for obstacle avoidance could
require them. The inclusions of images of arbitrary yaw movements into the dataset
would inflate its size to a great extent. This could lead to a deterioration of the accuracy
and increase the time-complexity of the k-NN algorithm. Instead, a ``derotation'' of the
camera image based on IMU data could be performed to align it with the underlying images of the dataset.

\section{CONCLUSION}
\label{sec:conclusion}

This paper presented an approach for lightweight indoor
localization of MAVs. We pursued an onboard design to foster
real-world use. The conducted experiments underline the applicability
of the system. Promising results were obtained for position estimates
and accurate landing in the indoor environment.

An important step in the approach is to shift computational effort to
a pre-flight phase. This provides the advantages of sophisticated
algorithms, without affecting performance during flight. The approach
can trade off speed with accuracy to use it on a wide range of models. The map
evaluation technique allows for predicting and improving the quality
of the approach.

\printbibliography

\end{document}